\documentclass{article}

\usepackage{PRIMEarxiv}

\usepackage[utf8]{inputenc} % allow utf-8 input
\usepackage[T1]{fontenc}    % use 8-bit T1 fonts
\usepackage{hyperref}       % hyperlinks
\usepackage{url}            % simple URL typesetting
\usepackage{booktabs}       % professional-quality tables
\usepackage{amsfonts}       % blackboard math symbols
\usepackage{nicefrac}       % compact symbols for 1/2, etc.
\usepackage{microtype}      % microtypography
\usepackage{lipsum}
\usepackage{fancyhdr}       % header
\usepackage{graphicx}       % graphics
\graphicspath{{media/}}     % organize your images and other figures under media/ folder

\usepackage{amsmath}
\usepackage{amssymb}
\usepackage{amsbsy}
\usepackage{amsthm}
\usepackage{algorithm} 
\usepackage{algpseudocode}
\usepackage{booktabs}
\usepackage{siunitx}
\usepackage{natbib}

\title{TESO: Tabu‐Enhanced Simulation Optimization for Noisy Black-Box Problems
}

\author{
  Bulent Soykan \\
  Institute for Simulation and Training \\
  University of Central Florida \\
  Orlando, FL, USA\\
  \texttt{Bulent.Soykan@ucf.edu} \\
   \And
  Sean Mondesire, Ghaith Rabadi \\
  School of Modeling, Simulation, and Training \\
  University of Central Florida \\
  Orlando, FL, USA\\
  \texttt{\{Sean.Mondesire, Ghaith.Rabadi\}@ucf.edu} \\
}

\begin{document}
\maketitle

\begin{abstract}
Simulation optimization (SO) is frequently challenged by noisy evaluations, high computational costs, and complex, multimodal search landscapes. This paper introduces Tabu-Enhanced Simulation Optimization (TESO), a novel metaheuristic framework integrating adaptive search with memory-based strategies. TESO leverages a short-term Tabu List to prevent cycling and encourage diversification, and a long-term Elite Memory to guide intensification by perturbing high-performing solutions. An aspiration criterion allows overriding tabu restrictions for exceptional candidates. This combination facilitates a dynamic balance between exploration and exploitation in stochastic environments. We demonstrate TESO's effectiveness and reliability using an queue optimization problem, showing improved performance compared to benchmarks and validating the contribution of its memory components. Source code and data are available at: \href{https://github.com/bulentsoykan/TESO}{github.com/bulentsoykan/TESO}.
\end{abstract}

% keywords can be removed

\section{Introduction}
\label{sec:intro}

Simulation Optimization (SO) is now widely recognized as a powerful methodology for decision-making across a diverse spectrum of fields, including engineering design \cite{deng2007simulation}, supply chain management \cite{jung2004simulation}, healthcare \cite{lal2015simulation} and increasingly, the tuning of hyperparameters in complex machine learning models \cite{amaran2016simulation}. The prevalence of SO stems from its power to identify optimal parameter settings or system configurations by directly interacting with simulation models, which often represent the only viable way to capture the intricacies and stochastic nature of real-world systems when analytical models are intractable \cite{10408315}.

However, the practical application of SO is frequently confronted by significant challenges. Many high-fidelity simulations are \textit{computationally expensive}, demanding substantial time or resources for each evaluation. These simulations are often treated as \textit{black-box} functions, meaning their internal analytical structure is either unknown or too complex to be exploited directly for optimization, thus preventing the direct calculation of derivatives \cite{TEKIN01112004}. Also, the outputs derived from \textit{stochastic simulations} are inherently noisy due to the underlying random components within the model. This noise complicates the assessment of a candidate solution's true performance, making it difficult to reliably distinguish genuine improvements from mere statistical fluctuations based on a limited number of simulation runs \cite{717096}. In addition, the search landscapes encountered in SO problems are frequently highly complex. They can be high-dimensional, contain numerous local optima (i.e., be \textit{multimodal}), and lack readily available or meaningful gradient information. This complexity renders standard gradient-based optimization techniques ineffective and necessitates the use of algorithms capable of global search without relying on derivative information \cite{figueira2014hybrid}.

A variety of approaches have been developed to tackle the challenges inherent in SO. Among the most prominent are Surrogate-Based Optimization (SBO) methods, which aim to mitigate the high computational cost of simulations by constructing cheaper-to-evaluate approximation models, or surrogates \cite{hong2021surrogate}. Gaussian Processes (GPs), often employed within a Bayesian Optimization (BO) framework, are frequently used for this purpose due to their ability to capture uncertainty in the approximation \cite{9004680}. However, constructing and tuning effective surrogate models, especially for high-dimensional or highly complex functions, can be computationally intensive itself and sensitive to the choice of kernel functions and hyperparameters \cite{fan2024review}.

Another significant class of methods, particularly relevant when comparing a finite number of system designs, falls under the umbrella of Ranking and Selection (R\&S) procedures \cite{boesel2003using}. These statistical methods focus on efficiently allocating simulation budget to identify the best among a discrete set of alternatives with a certain level of statistical confidence \cite{hong2021review}. While powerful in their domain, R\&S procedures are primarily suited for problems with discrete or a relatively small number of candidate solutions and are less directly applicable to the continuous or high-dimensional search spaces often encountered in optimization.

Metaheuristics represent a third major category, offering general-purpose stochastic search strategies applicable to complex black-box problems. Common examples used in SO include Genetic Algorithms (GA) \cite{de2015development}, Simulated Annealing (SA) \cite{haddock1992simulation}, and Particle Swarm Optimization (PSO) \cite{kuo2011simulation}. These methods often excel at global exploration but can suffer from premature convergence to local optima, particularly in the presence of significant simulation noise \cite{brahim2024metaheuristic}. Also, their performance can be highly sensitive to the choice of algorithmic hyperparameters (e.g., population size, mutation rates, cooling schedules), often requiring substantial tuning effort, which can itself become an optimization problem \cite{10408315}. Distinct from these, Tabu Search (TS) is a metaheuristic optimization strategy characterized by its intelligent use of memory structures \cite{glover1990tabu}. Central to TS is the concept of a \textit{Tabu List}, which records recently visited solutions or moves, temporarily forbidding them to prevent the search from cycling and to encourage exploration of unvisited areas of the solution space. This memory-based mechanism is specifically designed to help search processes escape the confines of local optima, a common pitfall for many other heuristic methods. The integration of such memory-based strategies offers a promising direction for enhancing the robustness and exploratory capabilities of SO algorithms.

Addressing the limitations of existing approaches and benefiting from the strengths of memory-based heuristics, this paper introduces Tabu-Enhanced Simulation Optimization (TESO), a novel algorithm specifically designed for complex and noisy SO problems. The core idea behind TESO is to dynamically balance the fundamental trade-off between \textit{exploration}, which involves discovering potentially new and promising regions of the vast solution space, and \textit{exploitation}, which focuses on carefully refining solutions within regions already identified as high-performing.  This research is motivated by the central research question: \textit{Can the integration of TS principles, Elite Memory structures, and adaptive perturbation strategies within a unified SO framework effectively balance exploration and exploitation to reliably converge towards high-quality solutions, particularly when dealing with noisy, potentially multimodal, and computationally expensive simulation environments?}

While TS and elite set strategies are known in deterministic optimization, their application to stochastic SO presents unique challenges. Standard TS assumes deterministic objective evaluations, making concepts like 'best move' and 'aspiration' straightforward. In SO, these must be redefined to operate on noisy estimates. TESO’s novelty is not in inventing these memory structures, but in adapting and integrating them into a cohesive framework for the SO context. Specifically, it uses a dual-memory approach which is explicitly designed to balance the search process under the uncertainty inherent to simulation.

To address the research questions, this paper makes the following primary contributions to the field of SO:
(i) The proposal of TESO, a \textit{novel SO algorithm} that synergistically integrates Tabu Search principles (tabu list, aspiration criterion) with Elite Memory and adaptive noise control, specifically tailored for noisy, complex black-box simulation environments.
(ii) An empirical \textit{demonstration of TESO's effective balancing of exploration and exploitation}. We show how tabu-driven diversification and random sampling interact with elite-guided intensification and adaptive perturbation to navigate challenging stochastic landscapes, leading to improved solution quality and reliability.

The remainder of this paper is structured as follows. Section \ref{sec:background} provides a detailed review of relevant background concepts. Section \ref{sec:methodology} presents the methodology of the proposed TESO algorithm, elaborating on its components. Section \ref{sec:experiments} details the numerical experiments conducted to evaluate TESO's performance. Finally, Section \ref{sec:conclusion} concludes the paper by summarizing the key findings, acknowledging limitations, and suggesting potential directions for future research in this area.

\section{Background and Related Work}
\label{sec:background}

\subsection{Simulation Optimization (SO)}
\label{subsec:so_definition}

SO refers to the process of finding optimal input parameters or decision variables, denoted by a vector $x$, for a system whose performance is evaluated using a computer simulation model \cite{andradottir1998simulation}. The goal is typically to minimize or maximize an objective function, $J(x)$, which represents the expected value of some performance measure obtained from the simulation output, $f(x, \omega)$ \cite{gosavi2015simulation}. Formally, the problem can often be expressed as:
\begin{equation}
    \min_{x \in \mathcal{X}} \quad J(x) = \mathbb{E}_{\omega}[f(x, \omega)]
\end{equation}
or equivalently for maximization, where $x$ represents the vector of decision variables, $\mathcal{X}$ is the feasible set of decisions, $f(x, \omega)$ is the performance measure output from a single simulation run given input $x$ and realization $\omega$ of the random elements within the simulation, and $\mathbb{E}_{\omega}[\cdot]$ denotes the expectation taken over the probability space governing the stochastic components $\omega$. A key characteristic of many SO problems is that the simulation model acts as a \textit{black box}, meaning an explicit analytical expression for $J(x)$ or its gradients is often unavailable \cite{cao2024blackbox}.

The application of optimization techniques within this simulation context presents several distinct and significant challenges. First and foremost is the need for \textit{handling stochastic responses} \cite{hong2021surrogate}. Since the simulation output $f(x, \omega)$ for any given $x$ is a random variable due to the inherent randomness $\omega$ within the simulation model, a single simulation run provides only a noisy estimate of the true expected performance $J(x)$. Relying on single, noisy observations can lead to incorrect assessments of solution quality and poor optimization decisions. Accurately estimating $J(x)$ typically requires multiple simulation replications for each candidate solution $x$, which directly impacts the overall computational effort. This inherent noise complicates tasks such as estimating gradients (if applicable), comparing candidate solutions reliably, and determining convergence.

Also, SO problems are often constrained by \textit{computational budget limitations} \cite{chen2011stochastic}. High-fidelity simulations representing complex real-world systems can be computationally intensive, requiring significant time or resources for each execution. Consequently, the total number of simulation runs (across all candidate solutions and all replications per solution) that can be performed is often severely limited. This budget constraint forces a careful allocation of computational effort and exacerbates the challenge posed by stochastic responses, as obtaining highly precise estimates of $J(x)$ for many different $x$ values may be infeasible.

In addition, SO algorithms must effectively balance the fundamental \textit{exploration versus exploitation dilemma} \cite{amaran2016simulation}. Given the limited budget and noisy feedback, the algorithm must decide whether to spend resources investigating new, unexplored regions of the decision space $\mathcal{X}$ (\textit{exploration}) in the hope of discovering globally superior solutions, or to focus resources on refining solutions within regions already known to yield good performance (\textit{exploitation}). Over-emphasizing exploration can waste valuable computational budget on evaluating unpromising areas, while over-emphasizing exploitation risks premature convergence to a locally optimal solution that may be significantly inferior to the true global optimum. Achieving an effective balance between these two competing objectives is critical for the success of any SO algorithm, particularly when dealing with the complex, potentially multimodal landscapes frequently encountered in simulation-based problems \cite{nsiye2024micro,soykanWintersim2024}. TESO is designed specifically to address these intertwined challenges.

\subsection{Metaheuristics in Simulation Optimization}
\label{subsec:metaheuristics}

Given the black-box nature and the frequent absence of readily available gradient information that characterize many SO problems, metaheuristic algorithms offer a compelling and widely adopted set of tools \cite{amaran2016simulation}. A primary strength of these population-based or trajectory-based methods lies in their inherent ability to perform a \textit{global search} across the decision space, rather than being confined to local improvements like gradient-based methods \cite{Soykan2022}. They operate directly on candidate solutions and their observed (potentially noisy) objective function values, bypassing the need for derivative information or strong assumptions about the objective function's structure (e.g., convexity). This makes them particularly suitable for solving the complex, potentially multimodal, and rugged landscapes often characteristic of simulation models \cite{thengvall2025measuring}.

However, the effective application of metaheuristics in SO is not without its difficulties. A significant challenge lies in \textit{parameter tuning}; algorithms like GA (requiring settings for mutation and crossover rates, population size, selection mechanisms), SA (requiring an effective cooling schedule), and PSO (requiring inertia weights, cognitive and social parameters) possess numerous control parameters. The optimal settings for these parameters can be highly problem-dependent and non-trivial to determine \textit{a priori}, often requiring extensive preliminary experimentation or specific expertise \cite{10408315}. Also, most metaheuristics are stochastic in nature and often lack formal \textit{convergence guarantees} to the global optimum, especially within a finite computational budget typically available in SO. While some theoretical convergence results exist under specific assumptions (e.g., for SA under infinitely slow cooling), practical guarantees of finding the true optimum in a finite number of steps are rare.

Despite their global search capabilities, these algorithms can still suffer from \textit{premature convergence}, becoming trapped in sub-optimal regions of the search space. The inherent \textit{stochasticity} of simulation outputs adds another layer of complexity; noise can obscure the true underlying landscape, making fitness or objective evaluations unreliable and potentially misleading the search direction, which can exacerbate issues like premature convergence or difficulty in discerning truly superior solutions. Despite these weaknesses, metaheuristics remain a popular and often effective approach for SO due to their flexibility, general applicability to challenging black-box problems, and potential for finding high-quality solutions when gradient-based or analytical methods are infeasible. TESO aims to retain the exploratory strengths while mitigating some weaknesses, particularly premature convergence, through its specific memory structures.

\subsection{Memory-Based Heuristics: Tabu Search and Elite Memory}  
\label{subsec:memory_heuristics_ts}  

TS is a prominent metaheuristic that exemplifies the strategic use of adaptive memory, particularly for escaping local optima in complex search spaces \cite{glover1990tabu}. Its core mechanism is the \textit{Tabu List}, a short-term memory that records recent moves or solution attributes, temporarily forbidding their reversal for a set duration (Tabu Tenure). This prevents cycling and promotes diversification by forcing the search into new areas. TS iteratively explores solution neighborhoods, selecting the best permissible move (non-tabu, or tabu but meeting an \textit{aspiration criterion}, such as finding a new global best) even if it results in a temporary worsening of the objective function. This non-improving move capability, combined with longer-term strategies for exploitation and exploration, has made TS highly effective \cite{Soykan2016}.

The broader concept of leveraging memory extends beyond TS, with many advanced heuristics incorporating mechanisms to learn from search history. A common and powerful form is the use of long-term memory to store a collection of \textit{elite solutions}—the best candidates found so far. This elite set serves as a basis for exploitation. Other forms of memory might track historical attribute frequencies or operator performance to adaptively bias the search. By incorporating memory, whether the short-term restrictive memory of a Tabu List or the long-term guiding memory of an elite set, heuristics can achieve a more effective balance between exploration and exploitation, leading to faster convergence and increased robustness, particularly in the noisy and complex landscapes characteristic of SO \cite{yu2023optimal}.

\section{TESO: Algorithm Methodology}
\label{sec:methodology}

\subsection{Framework Overview}
\label{subsec:framework_overview}

TESO is positioned as a direct search metaheuristic specifically adapted for noisy, black-box SO. It improves upon standard metaheuristics by integrating adaptive memory structures inspired by Tabu Search (TS) to better navigate complex, uncertain landscapes. TESO accommodates simulation noise by operating on mean estimates derived from multiple replications. Key adaptations include: a \textit{Tabu List} storing recent candidate representations to prevent re-evaluation and encourage diversification despite noise; a stochastic \textit{aspiration criterion} allowing tabu candidates to be accepted if their estimated performance surpasses the current best; and an \textit{Elite Memory} storing top-performing candidates based on their mean estimates, which provides robust starting points for intensification via perturbation.

When contrasted with \textit{standard TS}, TESO differs significantly in its neighborhood exploration and evaluation. Standard TS often relies on explicitly defined, discrete neighborhood structures (e.g., adjacent swaps, bit flips). TESO, on the other hand, employs an implicit neighborhood structure defined by adaptive random perturbation around elite solutions. The search step is probabilistic rather than a deterministic evaluation of all neighbors. Also, TESO evaluates candidates using multiple simulation replications to handle noise, whereas standard TS typically assumes deterministic function evaluations. This distinction is critical. In a deterministic setting, a move is unambiguously improving or not. In TESO, all comparisons are based on mean estimates from multiple replications, which are themselves random variables. Therefore, the aspiration criterion and improvement checks (detailed later) are inherently stochastic, a key adaptation for the SO domain.

Compared to \textit{other metaheuristics like GA, SA, or PSO} commonly used in SO, TESO's explicit use of \textit{both} short-term restrictive memory (Tabu List) and long-term guiding memory (Elite Memory) is a key differentiator. While GAs maintain diversity through populations and PSO uses particle/global bests (a form of Elite Memory), they typically lack the systematic short-term avoidance mechanism provided by a Tabu List. Basic SA is generally memoryless.  Also, TESO should be clearly distinguished from \textit{SBO and BO} approaches. SBO/BO methods construct an explicit statistical model (the surrogate, e.g., a Gaussian Process) of the underlying simulation response. They use this surrogate, along with an acquisition function (like Expected Improvement or Upper Confidence Bound), to intelligently select the next point(s) to evaluate with the expensive simulation, aiming to improve the surrogate model and find the optimum efficiently. TESO, in contrast, is a \textit{direct search} method. While it uses memory, it does not build an explicit global model of the objective function; it operates directly on the (replicated) outputs of the simulation model itself to guide its search trajectory.

\subsection{Algorithm Methodology}
\label{subsec:teso_methodology}  

TESO algorithm (Algorithm \ref{alg:TESO}) begins by initializing essential components (line 2): the Tabu List ($\mathcal{T}$) and Elite Memory ($\mathcal{E}$) are cleared, the best objective value found so far ($f_{\text{best}}$) is set to an appropriate initial value (e.g., infinity for minimization), the adaptive noise level ($\eta$) is set to its starting value ($\eta_{\text{init}}$), and iteration counters ($t$, $\Delta t$) are initialized. The core logic resides within a loop iterating up to the total trial budget $T$ (line 4). Each iteration commences with \textit{Candidate Generation} (lines 5-9). During the initial $n_{\text{init}}$ iterations or with a small probability $p_{\text{div}}$ thereafter, the algorithm performs diversification by generating a candidate $x^{(t)}$ randomly (line 6). Otherwise, it engages in intensification: an elite solution $x_e$ is selected from $\mathcal{E}$, and a new candidate $x^{(t)}$ is created by perturbing $x_e$ using the current noise level $\eta$ (line 8). A fallback to random generation occurs if $\mathcal{E}$ is empty. A unique, hashable representation $h^{(t)}$ of the generated candidate is then created (line 9). Since our test problem uses a continuous decision variable, this representation can be created by discretizing the variable into bins or by applying a hash function to its string representation. 

Before proceeding to potentially costly evaluation, the candidate undergoes a \textit{Tabu Check and Aspiration} step (lines 11-14). If the candidate's representation $h^{(t)}$ is found in the Tabu List $\mathcal{T}$ \textit{and} it does not satisfy the predefined aspiration criterion (i.e., it is not expected to significantly improve upon $f_{\text{best}}$), the candidate is discarded, and the algorithm proceeds directly to the next iteration via the `continue` statement (line 13). If the candidate $x^{(t)}$ is not tabu or satisfies the aspiration criterion, it is subjected to \textit{Stochastic Evaluation} (line 16). This involves running $n_{\text{rep}}$ independent simulation replications using $x^{(t)}$ as input. The mean performance $\mu^{(t)}$ and its standard deviation $\sigma^{(t)}$ are calculated from these replications, providing a statistically robust estimate of the candidate's objective value. Note that while the standard deviation$\sigma^{(t)}$ is calculated, it is not directly used in the decision logic of the current TESO implementation. It is recorded for analysis and could be used in more advanced versions of the algorithm, for instance, to dynamically adjust the number of replications or to inform a risk-based aspiration criterion.

Following evaluation, the algorithm performs \textit{Updates to the Best Solution and Memory Structures} (lines 18-22). It checks if the evaluated mean $\mu^{(t)}$ constitutes an improvement over the current $f_{\text{best}}$. If an improvement is found, $f_{\text{best}}$ and the corresponding best solution $x_{\text{best}}$ are updated, and the no-improvement counter $\Delta t$ is reset (line 20). If no improvement occurred (and the algorithm is past the initialization phase), $\Delta t$ is incremented (line 21). Regardless of improvement, the candidate's representation $h^{(t)}$ is added to the Tabu List $\mathcal{T}$, and the candidate-performance pair $(x^{(t)}, \mu^{(t)})$ is added to the Elite Memory $\mathcal{E}$, maintaining their respective capacity limits (line 22). Within the loop, \textit{Adaptive Noise Control and Termination Check} occur (lines 24-25). The noise level $\eta$ used for perturbation is updated according to a predefined schedule (e.g., decaying over iterations). The algorithm then checks if the termination criteria are met: either the total trial budget $T$ is exhausted (implicit in the `For` loop) or the number of iterations without improvement $\Delta t$ has reached the maximum allowed $\Delta t_{\text{max}}$. If the latter occurs, the loop terminates early via the `break` statement (line 25). Once the loop finishes, the algorithm returns the best solution $x_{\text{best}}$ and its corresponding estimated objective value $f_{\text{best}}$ (line 27).

\begin{algorithm}[h] 
\caption{TESO Algorithm}
\label{alg:TESO}
\footnotesize
\begin{algorithmic}[1]
\Require $f(x,\omega)$, $\mathcal{X}$, $T$, $n_{\text{init}}$, $n_{\text{rep}}$, $\eta_{\text{init}} \rightarrow \eta_{\text{final}}$, $C_{\mathcal{T}}, C_{\mathcal{E}}$, $p_{\text{div}}$, $\Delta t_{\text{max}}$, direction.
\Ensure Best solution $x_{\text{best}}$, Best objective $f_{\text{best}}$.

\State Init $\mathcal{T}, \mathcal{E}, f_{\text{best}}, \eta \gets \eta_{\text{init}}, \Delta t \gets 0, x_{\text{best}} \gets \text{null}$. \Comment{Initialize memory, best value, noise, counters}

\For{$t = 1$ to $T$}
    \State \Comment{Generate Candidate $x^{(t)}$}
    \If{$t \le n_{\text{init}}$ \textbf{or} Random() $< p_{\text{div}}$}
        \State $x^{(t)} \gets \text{RandCand}(\mathcal{X})$ \Comment{Diversify}
    \Else
        \State $x_e \gets \text{SelectElite}(\mathcal{E})$; $x^{(t)} \gets \text{Perturb}(x_e \text{ or RandCand()}, \eta)$ \Comment{Intensify}
    \EndIf
    \State $h^{(t)} \gets \text{Represent}(x^{(t)})$

    \State \Comment{Check Tabu \& Aspiration}
    \If{$h^{(t)} \in \mathcal{T}$ \textbf{and not} AspirCritMet($x^{(t)}$, $f_{\text{best}}$, direction)}
         \State \textbf{continue} \Comment{Skip tabu, non-aspirated}
    \EndIf

    \State \Comment{Evaluate Candidate}
    \State $(\mu^{(t)}, \sigma^{(t)}) \gets \text{Evaluate}(x^{(t)}, f, n_{\text{rep}})$ \Comment{Run $n_{rep}$ sims, get mean/std}

    \State \Comment{Update Best \& Memory}
    \If{IsImprovement($\mu^{(t)}$, $f_{\text{best}}$, direction)}
        \State $f_{\text{best}} \gets \mu^{(t)}$; $x_{\text{best}} \gets x^{(t)}$; $\Delta t \gets 0$
    \ElsIf{$t > n_{\text{init}}$} $\Delta t \gets \Delta t + 1$
    \EndIf
    \State Add $h^{(t)}$ to $\mathcal{T}$; Add $(x^{(t)}, \mu^{(t)})$ to $\mathcal{E}$ \Comment{Update memories}

    \State \Comment{Adapt Noise \& Check Termination}
    \State $\eta \gets \text{UpdateNoise}(t, T, \eta_{\text{init}}, \eta_{\text{final}})$
    \If{$\Delta t \ge \Delta t_{\text{max}}$} \textbf{break} \Comment{Stop if no improvement} \EndIf
\EndFor
\State \textbf{Return:} $x_{\text{best}}, f_{\text{best}}$.
\end{algorithmic}
\end{algorithm}

\begin{figure}
    \centering
    \includegraphics[width=1\linewidth]{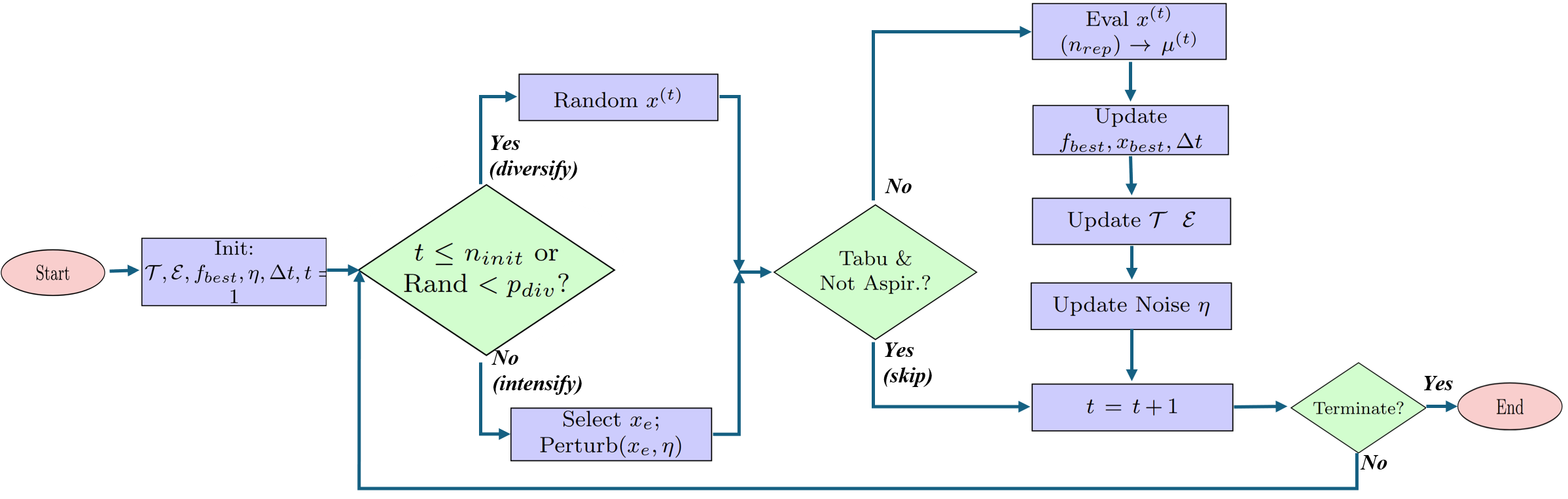}
    \caption{Flow Diagram of the TESO Algorithm, illustrating the iterative cycle including candidate generation, tabu/aspiration filtering, evaluation, and memory updates.}
    \label{fig:teso_flowchart}
\end{figure}

Figure \ref{fig:teso_flowchart} provides a visual summary of the TESO methodology detailed in Algorithm \ref{alg:TESO}. The diagram illustrates the flow starting from initialization, proceeding through the main iterative loop. Key stages depicted include the adaptive candidate generation (switching between diversification via random sampling and intensification via elite perturbation), the critical filtering step based on the Tabu List ($\mathcal{T}$) and aspiration criteria, the stochastic evaluation using multiple replications, and the subsequent updates to the best solution, both memory structures ($\mathcal{T}$, $\mathcal{E}$), and control parameters like noise level ($\eta$) before checking termination conditions.

\section{Numerical Experiments}
\label{sec:experiments}

\subsection{Experimental Objectives}
\label{subsec:exp_objectives}

The numerical experiments herein aim to empirically validate the proposed TESO algorithm's effectiveness for noisy SO problems. Our goals are fourfold: first, to demonstrate TESO's capability to converge reliably to high-quality solutions within typical budget constraints despite stochastic evaluations; second, to quantify the impact of its core memory components (Tabu List, Elite Memory) by comparing the full algorithm against ablation variants lacking one of these structures; third, to benchmark TESO's final solution quality, convergence speed, and reliability against baseline methods; and fourth, to analyze how TESO's mechanisms dynamically balance exploration and exploitation during the search. These experiments collectively seek to establish TESO as an effective framework for challenging SO problems.

\subsection{Test Problem: M/M/k Queue Optimization}
\label{subsec:test_problem}

To evaluate the performance of the TESO algorithm on a \textit{multiple-server M/M/k queueing system}. This system is characterized by Poisson arrivals (M), exponentially distributed service times (M) shared across $k$ identical parallel servers (k), and typically assumes an infinite buffer capacity operating under a First-In, First-Out (FIFO) discipline. The key parameters governing its behavior are the mean arrival rate, $\lambda$, the number of servers, $k$, and the mean service rate, $\mu$, for \textit{each individual server}. For stability, the system requires the total service capacity to exceed the arrival rate, i.e., $k \cdot \mu > \lambda$.

In this optimization context, the \textit{decision variable}, $x$, remains the \textit{service rate $\mu$} provided by each of the $k$ identical servers. We assume this rate can be controlled within a predefined feasible range $[\mu_{min}, \mu_{max}]$, where $k \cdot \mu_{min} > \lambda$ must hold for stability. The \textit{objective function}, $J(\mu)$, aims to minimize a combination of operational performance (customer waiting time) and the cost associated with providing service capacity. Specifically, we seek to minimize the sum of the expected steady-state average sojourn time ($W$) experienced by customers (time spent waiting in queue plus time in service) and a cost penalty proportional to the square of the individual server service rate, aggregated across all servers. The quadratic cost reflects the increasing marginal expense of providing faster service per server. Therefore, the objective function to be minimized is:
\begin{equation}
    J(\mu) = \mathbb{E}[W(k, \mu, \omega)] + C \cdot k \cdot \mu^2
    \label{eq:mmk_objective} % Changed label
\end{equation}
where $\mathbb{E}[W(k, \mu, \omega)]$ is the true expected average sojourn time when $k$ servers operate at rate $\mu$, and $C$ is a positive cost coefficient applied per server. For our experiments, we set the system parameters to create a non-trivial scenario: \textit{Arrival Rate:} $\lambda = 2.5$ customers per unit time, \textit{Number of Servers:} $k = 3$, \textit{Cost Coefficient:} $C = 0.5$, \textit{Feasible Service Rate Range:} The stability condition is $3\mu > 2.5$, or $\mu > 0.833...$. We define the feasible range for the individual server service rate as $\mu \in [1.0, 4.0]$.

This M/M/3 objective function retains the key characteristics relevant for testing TESO:

    \noindent \textit{Stochastic / Noisy Evaluation:} While complex analytical formulas exist for M/M/k steady-state probabilities (Erlang C formula), directly calculating the mean sojourn time $\mathbb{E}[W(k, \mu, \omega)]$ can be cumbersome or impractical within an optimization loop. It is typically estimated by running a discrete-event simulation of the M/M/3 system and averaging observed sojourn times. Random arrival and service processes ($\omega$) ensure each simulation yields a noisy estimate $\hat{W}(k, \mu, \omega)$, making the evaluated objective $\hat{J}(\mu) = \frac{1}{n_{rep}}\sum_{j=1}^{n_{rep}} \hat{W}(k, \mu, \omega_j) + C \cdot k \cdot \mu^2$ inherently stochastic.
    
    \noindent \textit{Black-Box Nature (from Optimizer's Perspective):} TESO treats the simulation providing $\hat{W}(k, \mu, \omega)$ as a black box, using only the input $\mu$ and the noisy output $\hat{J}(\mu)$.
    
    \noindent \textit{Complex Landscape:} Although the underlying trade-off between reducing waiting time (higher $\mu$) and increasing cost (higher $\mu$) likely results in a generally convex-\textit{like} shape for $J(\mu)$. Also, the noise present in the simulation-based evaluation creates a rugged landscape for the optimizer, posing a significant challenge for convergence and requiring robust handling of noisy feedback.

It is important to note that this M/M/3 problem is expected to be unimodal. While it effectively tests the algorithms' ability to handle noise and balance exploration and exploitation in a simple setting, it does not challenge their capacity to escape distinct local optima. The primary benefit of the Tabu List observed here will relate to preventing cycling in noisy regions rather than overcoming multimodality

\subsection{Benchmark Algorithms for Comparison}
\label{subsec:benchmarks}

We compare TESO's results against several benchmark algorithms to evaluate the performance and understand the contributions of the different components within the proposed TESO framework. These benchmarks are chosen to represent both a baseline level of performance and to facilitate an ablation study isolating the effects of TESO's core memory structures. The algorithms used for comparison are:

    \noindent \textbf{Pure Random Sampling (PRS):} This serves as the most basic benchmark. In PRS, candidate solutions $x^{(t)}$ are generated purely by sampling uniformly at random from the feasible decision space $\mathcal{X}$ for the entire duration of the optimization run ($T$ trials). No memory or adaptive search strategy is employed. PRS provides a baseline against which the added value of any intelligent search mechanism, including TESO's, can be measured. Its performance indicates the difficulty of finding good solutions simply by chance within the given budget.

    \noindent \textbf{TESO without Tabu List (TESO-noTabu):} This variant is identical to the full TESO algorithm (Algorithm \ref{alg:TESO}) except that the Tabu List mechanism is disabled. Comparing full TESO against TESO-noTabu allows us to directly assess the impact and benefit of the short-term tabu memory on preventing cycling, encouraging diversification, and potentially improving the final solution quality or convergence speed.

    \noindent \textbf{TESO without Elite Memory (TESO-noElite):} This variant mirrors the full TESO algorithm but removes the influence of the Elite Memory on candidate generation during the intensification phase. Instead of selecting an elite candidate $x_e$ from $\mathcal{E}$ and perturbing it, this variant might, for example, always perturb the single \textit{current} best-known solution $x_{\text{best}}$ or simply rely more heavily on the diversification mode (random generation). The comparison between full TESO and TESO-noElite aims to quantify the contribution of the long-term Elite Memory in effectively guiding the intensification process and exploiting promising regions identified during the search.

We have intentionally omitted comparisons to other common metaheuristics like SA or GA in this study. A fair comparison would require extensive hyperparameter tuning for each method, which itself constitutes a difficult optimization problem and could obscure the specific contributions of TESO's memory components. Our ablation study, therefore, provides a more controlled analysis of the direct impact of the Tabu List and Elite Memory, which is the primary focus of this paper.

\subsection{Experimental Design}
\label{subsec:exp_design}

To conduct a fair and comprehensive evaluation of TESO, we define a consistent experimental setup for optimizing the M/M/3 queue problem (\textbf{$k=3$, $\lambda=2.5$, $C=0.5$, $\mu \in [1.0, 4.0]$}). Each algorithm (TESO and benchmarks) is run for a total budget of $T=300$ candidate evaluations, with the first $n_{\text{init}}=20$ dedicated to random initialization. Candidate solutions are evaluated using $n_{\text{rep}}=30$ independent simulation replications to estimate the mean objective and its standard deviation. For statistical robustness, all results are averaged over $N_{\text{macro}}=30$ independent macro-replications. TESO utilizes specific parameter settings: a linearly decaying noise schedule for perturbations ($\eta_{init}=0.2$ to $\eta_{final}=0.01$), a Tabu List capacity $C_{\mathcal{T}}=15$, an Elite Memory capacity $C_{\mathcal{E}}=10$, a diversification probability $p_{\text{div}}=0.2$, and termination if no improvement occurs in $\Delta t_{\text{max}}=50$ iterations. The benchmark algorithms adapt this setup: \textit{PRS} omits memory and adaptive search; \textit{TESO-noTabu} disables the Tabu List ($C_{\mathcal{T}}=0$); and \textit{TESO-noElite} modifies intensification to not rely on the elite set $\mathcal{E}$. All algorithms operate on the same queue problem defined in Section \ref{subsec:test_problem}.

Performance is assessed using several key metrics across the 30 macro-replications. We report the average and standard deviation of the \textit{Best Mean Objective Value Found} ($f^*_{\text{best}}$) upon termination to gauge final solution quality and reliability. The stability of performance near the end of the search is measured by the \textit{Average Objective Value over the Last 50 Trials}. \textit{Convergence Plots} illustrate the progression of the average best-found solution over the trial budget, visually comparing speed and consistency (often including standard error bands). Finally, we record the average \textit{Computational Time} per macro-replication as a practical measure of efficiency. These metrics collectively allow for a thorough comparison of the effectiveness, reliability, and efficiency of the algorithms.

\subsection{Results and Discussion}
\label{subsec:results_discussion}

\begin{figure}
    \centering
    \includegraphics[width=0.8\linewidth]{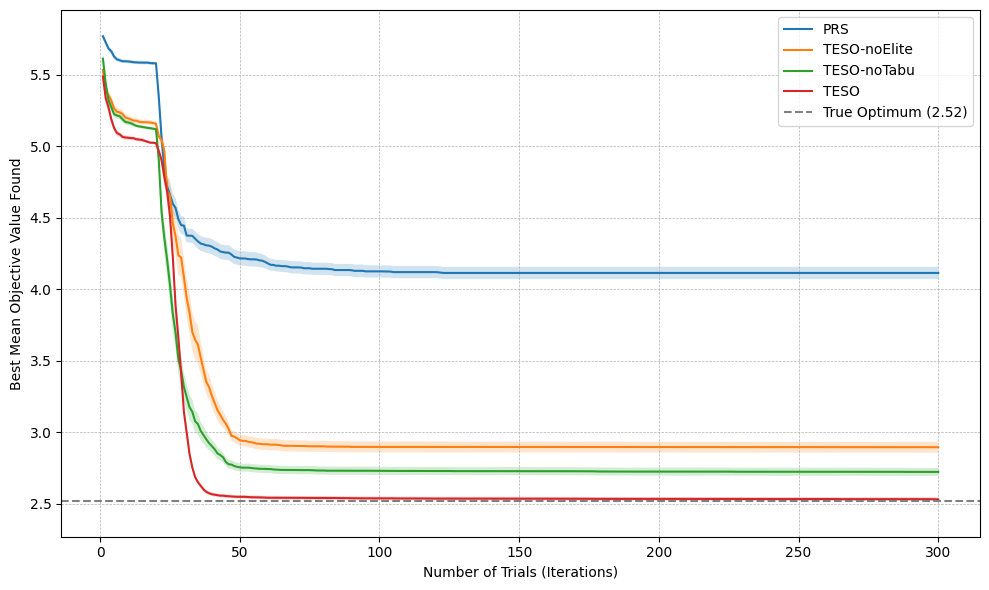}
    \caption{Convergence Plot for Queue Optimization}
    \label{fig:plot}
\end{figure}

The performance of the proposed TESO algorithm and the selected benchmarks are evaluated based on the  queue optimization problem described in Section \ref{subsec:test_problem}. The results, averaged over $N_{\text{macro}}=30$ independent macro-replications, are presented numerically in Table \ref{tab:results_summary} and visually through the convergence plots in Figure \ref{fig:plot}.

\paragraph{Comparison against Benchmarks}
As expected, PRS demonstrates the weakest performance, achieving the highest (worst) final best mean objective value (approximately 4.11) with the largest standard deviation (0.20), indicating low solution quality and poor reliability. The convergence plot for PRS shows slow improvement, confirming its inefficiency as a search strategy. In contrast, all TESO variants significantly outperform PRS. The full TESO algorithm achieves the best overall performance, converging to the lowest final best mean objective value (2.53), which is very close to the assumed true optimum (2.52). Furthermore, TESO exhibits the lowest standard deviation for the final best value (0.07) across the macro-replications, highlighting its superior reliability and consistency in finding high-quality solutions compared to all benchmarks. While TESO incurs a slightly higher average computational time compared to its ablation variants, this is justifiable given the significant improvement in solution quality and reliability. The convergence plot (Figure \ref{fig:plot}) confirms TESO's faster convergence towards better solutions compared to the other methods.

\paragraph{Impact of Memory Components (Ablation Study)}
Comparing the full TESO algorithm with its variants allows us to analyze the specific contributions of the Tabu List and Elite Memory.

\noindent\textbf{Effect of Tabu List:} Comparing TESO with TESO-noElite (which has Tabu but no Elite guidance for intensification) against PRS shows the baseline benefit of using memory. Comparing TESO (full) with TESO-noTabu reveals the advantage conferred by the tabu mechanism. TESO-noTabu achieves a respectable final objective value (2.72), substantially better than PRS, demonstrating the effectiveness of elite-guided intensification. However, it performs worse than the full TESO (2.53) and exhibits higher variability (std dev 0.16 vs 0.07). This suggests that without the Tabu List, the search, while guided by elite solutions, may indeed spend unnecessary effort revisiting recently explored areas or get temporarily stuck near elite candidates, hindering diversification and slowing convergence to the very best solutions. The Tabu List effectively mitigates this.

\noindent\textbf{Effect of Elite Memory:} Comparing TESO with TESO-noTabu (which has Elite but no Tabu) against PRS shows the benefit of intensification. Comparing TESO (full) with TESO-noElite highlights the role of guided intensification. TESO-noElite performs better than PRS but worse than both TESO-noTabu and the full TESO (final mean obj. 2.89, std dev 0.21). This indicates that while the Tabu List helps diversification, the lack of targeted intensification based on a pool of elite solutions (instead potentially perturbing only the single best or relying more on random jumps) makes the exploitation phase less efficient. The Elite Memory is crucial for effectively focusing the search and refining high-potential regions identified during exploration.
The results strongly suggest that both the short-term tabu memory and the long-term Elite Memory play vital and complementary roles in TESO's performance. The Tabu List enhances exploration and prevents stagnation, while the Elite Memory effectively guides exploitation, leading to a synergistic effect in the full TESO algorithm that outperforms either component used in isolation (within the TESO framework).

\paragraph{Exploration and Exploitation Balance}
The convergence plots provide insights into the exploration-exploitation dynamics. PRS represents pure exploration with no learning. TESO-noElite, lacking strong guidance for exploitation, shows slower convergence after the initial phase compared to methods using Elite Memory. TESO-noTabu shows faster initial convergence due to elite guidance but potentially plateaus earlier or at a slightly higher level than full TESO, possibly due to insufficient diversification caused by revisiting areas near elite solutions. The full TESO algorithm exhibits a desirable pattern: a period of exploration (similar slope to others initially, possibly slightly slower than TESO-noTabu if elite guidance is very strong early on), followed by a phase of rapid improvement (steeper slope) as intensification guided by Elite Memory takes over, and finally converging reliably to the best solution region with low variance. This behavior suggests that the combination of tabu diversification, elite intensification, and adaptive noise successfully manages the exploration-exploitation trade-off for this noisy SO problem.

\begin{table}[htbp] % Use htbp for better placement options
\centering
\footnotesize
\caption{Performance Comparison of TESO and Benchmark Algorithms.}
\label{tab:results_summary} % Changed label to avoid conflict if old table exists
% Using siunitx for better alignment (optional, requires \usepackage{siunitx})
\sisetup{round-mode=places, round-precision=2, table-format=1.2(2)} % Format: 1 digit before ., 2 after, (2) for std dev in parens
\begin{tabular}{l S[table-format=1.2] S[table-format=1.2] S[table-format=1.2] S[table-format=3.2]} % Adjust table-format as needed
% Simpler formatting without siunitx:
% \begin{tabular}{l c c c c}
\toprule
          & {Final Best} & {Final Best}    & {Avg Obj. } & {Avg Comp.} \\
Algorithm & {Mean Obj.}  & {Std Dev} & {(Mean $\pm$ Std Dev)}     & {Time (s)} \\
\midrule
PRS           & 4.11 & 0.23 & 4.11 $\pm$ 0.20  & 53.34 \\ 
TESO-noElite  & 2.89 & 0.21 & 2.89 $\pm$ 0.18  & 101.43 \\ 
TESO-noTabu   & 2.72 & 0.16 & 2.72 $\pm$ 0.13  & 113.52 \\ 
\textbf{TESO} &   \textbf{2.53} &   \textbf{0.07} &   2.53 $\pm$ 0.06  &   132.81  \\ 
\bottomrule
\end{tabular}
\end{table}

\section{Conclusion and Future Work}
\label{sec:conclusion}

This paper introduced TESO, a novel metaheuristic framework tailored for optimizing noisy, expensive black-box simulation models. \textit{TESO's methodology} combines adaptive random search with a short-term Tabu List to prevent cycling and a long-term Elite Memory to guide intensification around promising solutions, employing an aspiration criterion and adaptive noise control. Our \textit{key experimental results} on a queue optimization problem demonstrated TESO's effectiveness, showing consistent convergence to high-quality solutions with better reliability (lower variance) compared to PRS and ablation variants lacking either memory component. The study confirmed the synergistic benefit of integrating both tabu restrictions and elite guidance, leading to a \textit{successful balance between exploration and exploitation}. The primary \textit{contributions} include this novel integration of TS principles and Elite Memory for stochastic optimization, the empirical demonstration of improved performance, and a structured approach to managing the exploration-exploitation trade-off. However, we acknowledge \textit{limitations} such as validation on a single, unimodal problem class, potential sensitivity to TESO's internal parameters, and untested scalability in very high-dimensional settings. Furthermore, while we criticize other metaheuristics for their sensitivity to hyperparameters, TESO is not immune. Its performance depends on parameters and the noise decay schedule. A full sensitivity analysis was beyond the scope of this initial study but is a critical direction for future work. Finally, the algorithm's scalability in very high-dimensional settings remains untested.

Future work will focus on three key areas. First, we will test TESO's generalizability on a wider range of benchmark problems with higher dimensions and multimodality. Second, we aim to enhance the algorithm by developing adaptive mechanisms for its internal parameters and exploring hybridization with surrogate models. Finally, pursuing a theoretical analysis of its convergence properties would provide important formal grounding.

%Bibliography
\bibliographystyle{unsrt}  
\bibliography{references}

\end{document}